\documentclass[10pt,twocolumn,letterpaper]{article}

\usepackage[pagenumbers]{cvpr} 

\usepackage[dvipsnames]{xcolor}

\usepackage{multirow}

\definecolor{cvprblue}{rgb}{0.21,0.49,0.74}
\usepackage[pagebackref,breaklinks,colorlinks,citecolor=cvprblue]{hyperref}

\def\paperID{979} 
\def\confName{CVPR}
\def\confYear{2024}

\title{Interactive Humanoid: Online Full-Body Motion Reaction Synthesis \\with Social Affordance Canonicalization and Forecasting }

\author{
Yunze Liu \textsuperscript{1,3},
\quad Changxi Chen \textsuperscript{1},
\quad Li Yi\textsuperscript{1,2,3},\\
\textsuperscript{1} Tsinghua University,
\quad\textsuperscript{2} Shanghai Artificial Intelligence Laboratory,
\quad\textsuperscript{3} Shanghai Qi Zhi Institute\\
}

\begin{document}
\maketitle
\begin{abstract}

We focus on the human-humanoid interaction task optionally with an object. We propose a new task named online full-body motion reaction synthesis, which generates humanoid reactions based on the human actor's motions. The previous work only focuses on human interaction without objects and generates body reactions without hand. Besides, they also do not consider the task as an online setting, which means the inability to observe information beyond the current moment in practical situations. 
To support this task, we construct two datasets named HHI and CoChair and propose a unified method. Specifically, we propose to construct a social affordance representation. We first select a social affordance carrier and use SE(3)-Equivariant Neural Networks to learn the local frame for the carrier, then we canonicalize the social affordance. Besides, we propose a social affordance forecasting scheme to enable the reactor to predict based on the imagined future.
Experiments demonstrate that our approach can effectively generate high-quality reactions on HHI and CoChair. Furthermore, we also validate our method on existing human interaction datasets Interhuman and Chi3D. Website:  \href{https://yunzeliu.github.io/iHuman/}{https://yunzeliu.github.io/iHuman/} 
\vspace{-7mm}
\end{abstract}

\section{Introduction}
In various applications including VR/AR, games, and human-robot interaction, there is a strong demand for generating reactive humanoid characters or robots based on the actions of human actors. Such a reaction needs to occur in real-time, dynamically responding to the movements of the human actor. Furthermore, in many cases, these interactions involve objects (e.g., a human and a humanoid collaboratively carrying a chair) and call for an emphasis on the precise movements of humanoid hands in addition to the overall body motion. Addressing the challenge of synthesizing humanoid\footnote{In this paper, we use \textit{human} to denote real people initiating interactions and \textit{humanoid} to indicate the virtual character reacting in response.} reactions in these contexts can significantly enhance the social experience of humans interacting with humanoids and bring a whole new entertainment.

Previous research on humanoid motion synthesis has mainly focused on single humanoid movements~\cite{guo2020action2motion,xu2023actformer,athanasiou2022teach, zhang2022motiondiffuse, petrovich2022temos,dabral2023mofusion,jiang2023motiongpt} or interactions with objects~\cite{zheng2023cams, xu2023interdiff,zhang2023artigrasp,zhu2021toward,christen2022d,zhang2021manipnet},. Some recent studies~\cite{liang2023intergen,xu2023actformer}, have explored the synthesis of social interactions between two humanoids. However, these studies have limitations. Firstly, they primarily focus on the offline generation between two humanoids, which is not suitable for the asymmetric reaction synthesis setting where the humanoid continuously responds to the dynamic human actor in an online manner. Secondly, they overlook the fact that humans often interact through objects. Thirdly, these works do not consider synthesizing full-body motions involving both body and hand motions, which are crucial for various interactions such as handshakes or collaborations. Therefore, synthesizing full-body humanoid reactions online considering both human actors and the possible objects goes beyond the scope of existing works, presenting three major challenges: 1) representing complex motions of a human actor and optionally an object, 2) interpreting the human actor's intentions for prompt reactions by the humanoid, and 3) supporting detailed reactions involving both coarse-grained body movements and fine-grained hand movements.

\begin{figure}[t]
\includegraphics[width=\linewidth]{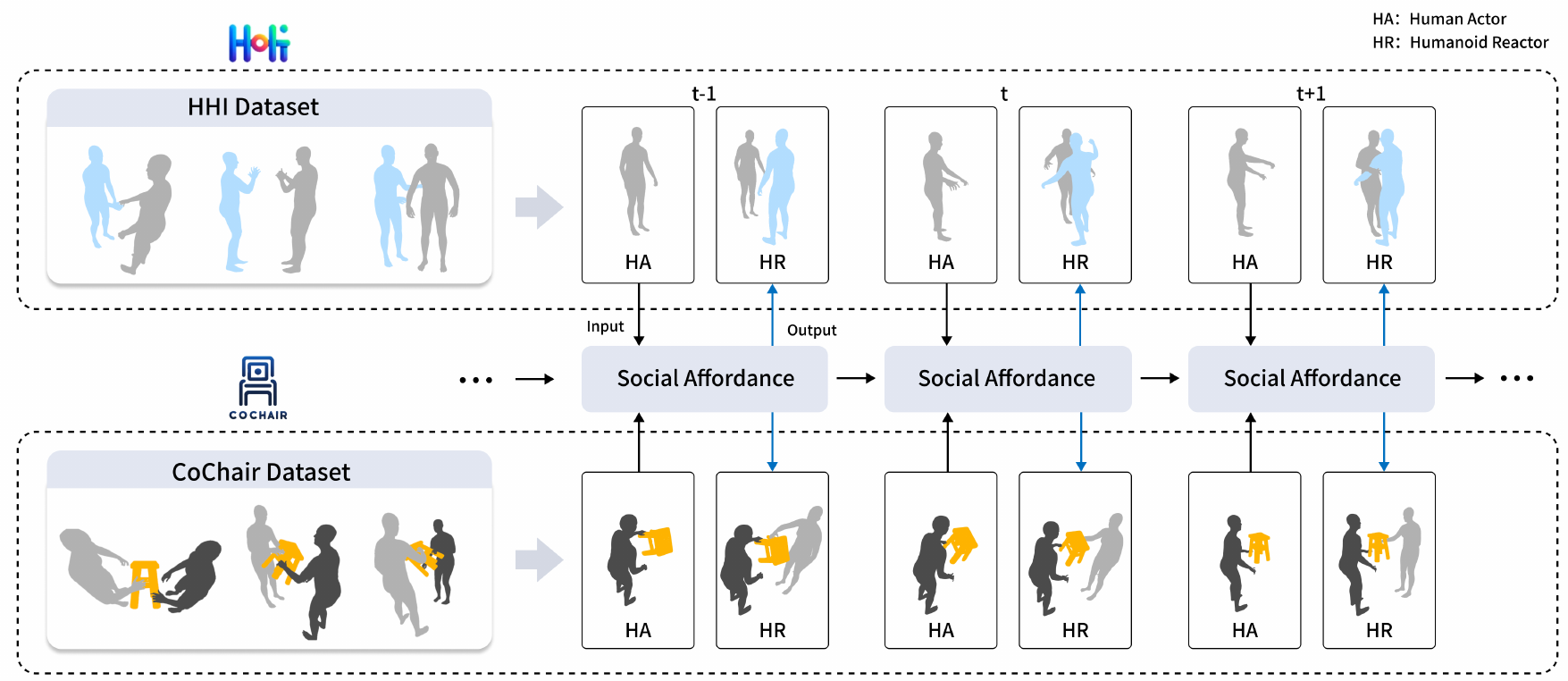}
    \centering
    \vspace{-5mm}
    \caption{We propose a new task named online full-body motion reaction synthesis optionally with an object. Left: we construct two datasets HHI and CoChair to support the task. Right: we propose Social Affordance Canonicalization and Forecasting technique to generate realistic reactions and establish benchmarks.}
    \label{fig:teaser}
    \vspace{-8mm}
\end{figure}

To address the challenges mentioned, we draw inspiration from affordance learning. Our approach involves encoding the motion of the human actor (possibly with an object) as a social affordance representation, capturing supported and expected social interactions at both the body and hand levels. Subsequently, we learn the humanoid's reaction based on the social affordance representation. To simplify the distribution of social affordance and facilitate learning, we employ a canonicalization technique. Furthermore, we introduce an online social affordance forecasting scheme to enable the humanoid to react promptly.

Specifically, our method handles a sequential input data stream comprising the human actor's pose at each time step. When objects are involved, the input also includes a stream of 6D-transforming 3D shapes. To unify the input with and without objects, we introduce the concept of ``affordance carrier'', which can be either the real object in human-object-humanoid interactions or just the reactor in a rest state in human-humanoid interaction. Centering around the affordance carrier, we propose a social affordance representation encompassing the actor's motion, the carrier's dynamic geometry, and the actor-carrier relationship up until each time step. And then we learn to predict the humanoid's online reaction through a 4D motion transformer. However, we find directly predicting the reaction based on the social affordance representation not satisfactory enough for two reasons. First, the actor's motion exhibits diverse patterns, complicating the social affordance and increasing the learning difficulty. To address this, we observe that the patterns within the actor's motion become more compact when viewed from the carrier's local coordinate system. Therefore, we present a social affordance canonicalization strategy enabled by equivariant local frame learning. Second, the humanoid reactor can only access present and past observations, limiting its social affordance to short-sighted information and hindering prompt reactions. To overcome this, we propose a social affordance forecasting scheme to enable the reactor to predict based on the imagined future.

To validate the effectiveness of our design and also to address the lack of large-scale full-body reaction-synthesis benchmarks, we have gathered two large-scale full-body social interaction datasets named HHI and CoChair. HHI covers a diverse range of human-human interactions with a clear actor and reactor while CoChair focuses on human-object-human interaction. Our method consistently outperforms previous methods in all metrics.
We also validate our method on existing datasets Interhuman and Chi3D.

The key contributions of this paper are threefold: i) we propose a new task named online full-body motion reaction synthesis optionally with an object and establish benchmarks;
ii) we construct two datasets HHI and CoChair to support the research on full-body reaction synthesis tasks;
iii) we propose a unified solution to reaction synthesis with or without objects by social affordance canonicalization and forecasting, significantly outperforming baselines.

\section{Related Work}
\noindent\textbf{Human Motion Generation.}
Human motion generation is to generate human motion conditioned on different signals. A line of works\cite{cervantes2022implicit, chen2023executing, guo2020action2motion, petrovich2021action,xu2023actformer,yan2019convolutional,athanasiou2022teach} propose to generate human motion conditioned on action label. Some works\cite{ahuja2019language2pose, zhang2022motiondiffuse, petrovich2022temos,guo2022generating,kim2023flame,dabral2023mofusion,jiang2023motiongpt} directly generate human motion conditioned on text description. There are also some works\cite{lee2019dancing,li2022danceformer,li2021ai} that generate human motion conditioned on music and speech. Recently, some works\cite{xu2023actformer,shafir2023human,liang2023intergen,starke2020local} have started to focus the human-human interaction synthesis. \cite{liang2023intergen} propose a new dataset with natural language descriptions and design a diffusion model to generate human-human interaction. However, this method cannot be directly applied to reaction synthesis because it uses a fixed CLIP\cite{radford2021learning} branch to extract text features. \cite{xu2023actformer} presents a GAN-based Transformer for action-conditioned motion generation. However, it cannot generate full-body motions and cannot handle the presence of objects. 

\noindent\textbf{Human Reaction Generation.}
 We focus on motion generation conditioned on another human motion. Human reaction generation is conditioned on the actor's motion and requires the reactor to provide a reasonable response, which is very important in the fields of VR/AR and humanoid robots. \cite{chopin2023interaction} propose a Transformer network with both temporal and spatial attention to generate reactions. \cite{baruah2023intent} propose to predict human intent in Human–Human interactions. 
However, they are only concerned with the generation of body motions and cannot generate hand motions. At the same time, they only focus on the interaction between humans and cannot generate reasonable reactions in the presence of objects. In addition, reaction synthesis should be in an online setting, meaning that the reactor cannot observe future information, which is also not discussed in previous work. There is no dataset providing full-body human-human interaction and human-object-human interaction with clear actor and reactor, so in this paper, we first construct two datasets and propose a novel method to generate realistic reactions.

\noindent\textbf{Human Motion Prediction}
Human motion prediction~\cite{ijcai2022p111, martinez2017human, li2018convolutional, tang2018long, corona2020context, hernandez2019human, mao2019learning, sofianos2021space, zhong2022spatio, sofianos2021space} is a traditional task widespread attention. 
A line of works predicts human motions in an encoding-decoding way~\cite{salzmann2022motron, lucas2022posegpt, blattmann2021behavior, dang2022diverse, xu22stars}. Some works carefully designed loss constraints~\cite{aliakbarian2020stochastic, gupta2018social, lee2017desire} to generate diversity and realistic human motions. 
Without multi-stage training,\cite{chen2023humanmac} propose a human motion diffusion model to predict human motion in a masked completion fashion. Besides, \cite{xu2023interdiff} propose to predict human motion with the object as an HOI sequence and use interaction diffusion and interaction correction to predict the future state of human and object. In this paper, we focus on human-human and human-object-human interactions. We propose to use a motion forecasting module to improve the ability of the reactor, thus relying on both human motion prediction and human-object interaction prediction methods. 

\begin{figure*}[t] 
	\centering  
	\includegraphics[width=1\textwidth]{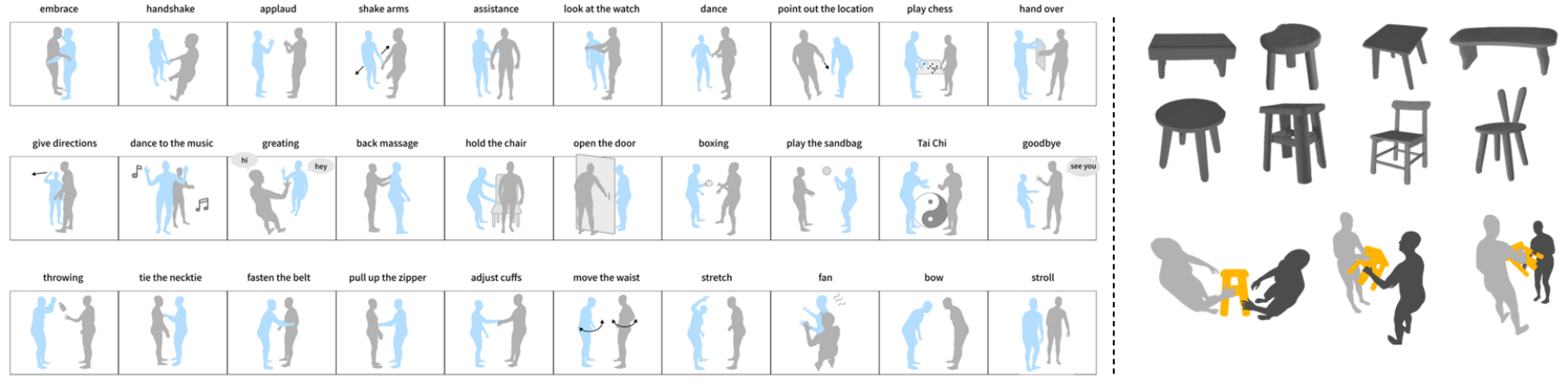} 
 	\vspace{-6mm} 
	\caption{We construct two datasets to support the research on reaction synthesis. HHI(left) is the first large-scale whole-body motion reaction dataset with clear action feedback. CoChair(right) is the first large-scale dataset for multi-human and object interaction} 
	\vspace{-8pt} 
    \label{fig:dataset} 
\end{figure*}

\section{Datasets}

\subsection{HHI: Human-Human Interaction }

The HHI dataset is a large-scale full-body motion reaction dataset with clear action feedback which includes 30 interactive categories, 10 pairs of human body types, and a total of 5,000 paired interaction sequences. 
The HHI dataset has three characteristics.
The first is \textit{multi-human whole-body interaction} including body and hand interactions which is not supported in previous datasets. We believe hands are essential in multi-human interactions and can convey rich information during handshake, hug, and handover.
The second is our dataset can distinguish \textit{distinct actors and reactors}. For example, during handshakes, pointing directions, greetings, handovers, etc., We can identify the initiator of the action, which can help us better define and evaluate this problem.
The third is our dataset has a \textit{rich diversity}, which includes the types of interactions and reactions. We not only include 30 types of interactions between two humans but also provide multiple reasonable reactions for the same action by the actor. For example, when someone greets you, you can nod in response, respond with one hand, or respond with both hands. This is also a natural feature that humans possess when reacting, but previous datasets have rarely focused on this point and discussed it.

\subsection{CoChair: Human-Object-Human Interaction}
CoChair is a large-scale dataset for multi-human and object interaction consisting of 8 different chairs, 5 interaction patterns, and 10 pairs of different skeletons, totaling 3000 sequences.
CoChair has two significant characteristics. CoChair is information asymmetry during collaboration. Each action has an actor/initiator (who knows the destination of the carrying) and a reactor (who does not know the destination), which can support our research on the reactor's behavior patterns. The second is the diversity of carrying patterns. The dataset includes five carrying patterns: one-hand fixed carrying, one-hand mobile carrying, two-hand fixed carrying two-hand mobile carrying, and two-hand flexible carrying.

\subsection{Dataset Construction.}
\noindent\textbf{Dataset Comparison.} HHI is the first large-scale dataset with diverse interactions for whole-body reaction synthesis. It not only provides motion capture of the whole body but also designs interaction with clear initiators. Compared to the challenging free-from interaction provided by Interhuman, our dataset has explicit categories of interactive actions, which facilitates the evaluation of generated results. Compared to datasets such as SBU, K3HI, and Chi3D which fully or partially use image-based methods to estimate human poses, our dataset is completely captured by motion capture devices and meticulously annotated by human experts, which can provide higher-quality motions.

CoChair is the first large-scale dataset for human-object-human collaborative carrying. It not only has clear motion initiators but also diverse object geometries and different carrying patterns. Compared to other datasets, we support a more challenging setting that involves not only human interaction but also the interaction between humans and objects. 

\begin{table}[th]
    \begin{center}
        \centering
        \resizebox{0.48\textwidth}{!}{
        \begin{tabular}{lccccccccccc}
            \toprule
            Dataset & Object &Full-body  & Actor\&Reactor  & Mocap  &Motions   &Verbs  & Duration      \\
            \midrule
                SBU\cite{yun2012two} & -& - &-   & - & 282    & 8       & 0.16h   \\
                K3HI\cite{baruah2020multimodal} &-  &- &-  & -  & 312    & 8     & 0.21h  \\
                NTU120\cite{liu2019ntu} &- &-   & - &-  & 739   & 26    & 0.47h \\	
                You2me\cite{ng2020you2me}&-    &-   & -  &-   & 42    & 4  & 1.4h \\
                Chi3D\cite{fieraru2020three}&- &-   &-   &$\sqrt{}\mkern-9mu{\smallsetminus}$  & 373    & 8     & 0.41h   \\
                InterHuman\cite{liang2023intergen}&-  &-   &- &$\checkmark$   & 6022    &5656    &6.56h       \\
            \midrule
            \textbf{HHI (Ours)} & -  & $\checkmark$    &  $\checkmark$  & $\checkmark$ & 5000  & 30 &  5.55h  
            \\
            \textbf{CoChair (Ours)} & $\checkmark$  & $\checkmark$    &  $\checkmark$  & $\checkmark$ & 3000  & 5 &   2.78h 
            \\
            \bottomrule
        \end{tabular}
        }
          \caption{\textbf{Dataset comparisons.} We compare our iHuman dataset with existing multi-human interaction datasets. \textbf{Object} refers to human-object-human interaction. \textbf{Whole-body} refers to whole-body motion capture. \textbf{Actor\&Reactor} refers to whether there is an obvious initiator of the action. \textbf{Motions} is the total number of motion clips. \textbf{Verbs} is the number of interaction categories. \textbf{Duration} refers to the total time of each dataset.}\label{tab:dataset}
        \vspace{-5mm}
    \end{center}
\end{table}


\noindent\textbf{Dataset Collection.}
We use 12 NOKOV motion capture cameras with 60fps for data collection. The cameras are arranged in a square formation on the ceiling, with four cameras on each side. For the body, each participant has to attach 53 markers. As for the hands, we attached 32 markers to the wrists, finger joints, and fingertips (16 markers per hand). For tracking objects during interactions, we affix 9 markers to each object. 
During the interaction, only the actor knows the task and we allow them to decide whether or not to share this information with the reactor. Then the actor decides when to start the carrying process. This ensures the authenticity of the reaction when it responds. The actor will initiate the motion first, and then the reactor will respond accordingly based on instinct and habit.

\noindent\textbf{Dataset annotation.}
We track 53 markers on the body and 16 markers on each hand.  Followed by AMASS\cite{mahmood2019amass}, we obtain SMPL-X\cite{pavlakos2019expressive} parameters for each frame through optimization. 
For each object, we mark the position of the attached marker on the scanned CAD model and calculate the transformation matrix.

\section{Method}
\begin{figure*}[t] 
	\centering  
	\includegraphics[width=1\textwidth]{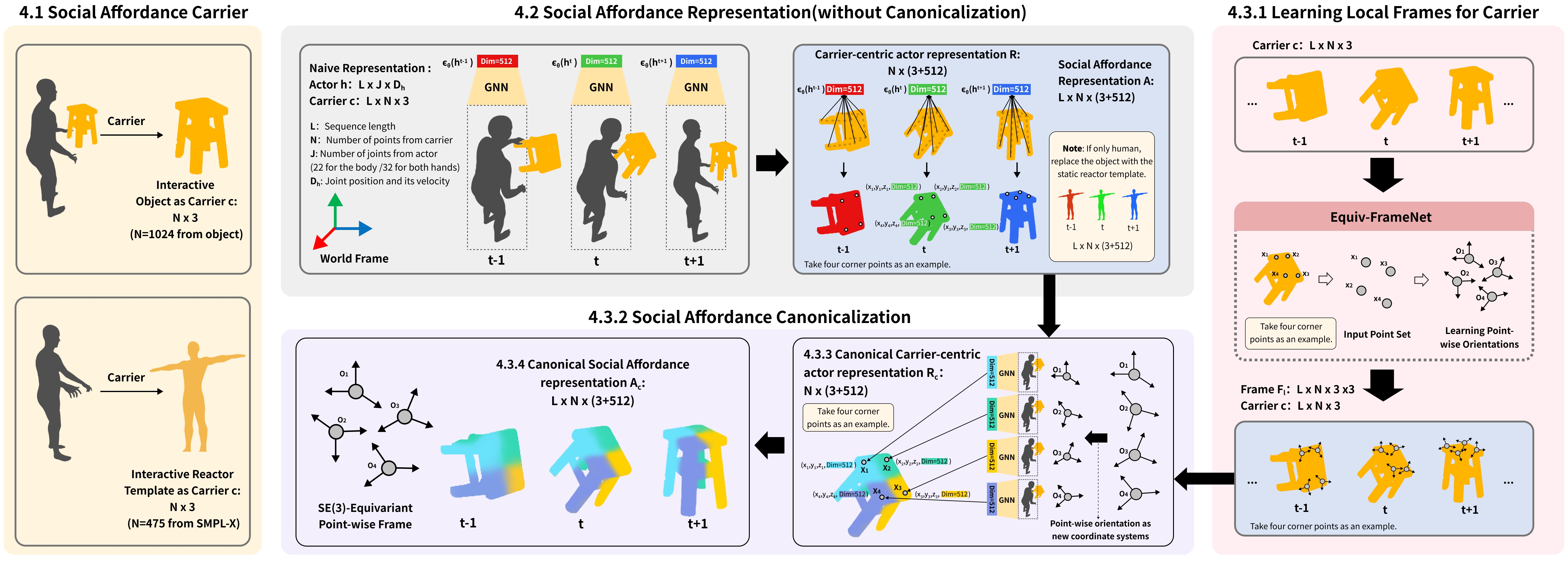} 
 	\vspace{-8mm} 
	\caption{Social Affordance Canonicalization. Given a sequence, we first select a social affordance carrier and build the carrier-centric representation. Then we can compute the social affordance representation. We propose to learn the local frame for carrier and canonicalize social affordance to simplify the distribution. Then a motion encoder and decoder are used to generate reactions.} 
	\vspace{-5mm} 
    \label{fig:canonical} 
\end{figure*} 

Our goal is to generate motions for the humanoid to interact or collaborate with the human in social scenarios or collaborative tasks, which requires the humanoid to not only understand the human's intentions and motions but also comprehend the state of the environment or objects.
In this Section, we elaborate on the method in detail. We first introduce the concept of social affordance carrier and carrier-centric representation. Then we introduce the social affordance canonicalization technique to simplify the distribution of social affordance and facilitate learning. We also introduce an online social affordance forecasting
scheme to enable the humanoid to react promptly. Finally, we show the 4D motion transformer and objective function for the entire framework.

\subsection{Social Affordance Carrier}
Social affordance carrier refers to the object or humanoid encoding social affordance information. When a human actor interacts with a humanoid, the human actor often comes into contact with the humanoid directly or indirectly. And when there is an object, human actors often come into direct contact with the object. In order to model the direct or potential contact information in the interaction, we need to choose a carrier to simultaneously represent the human actor, the carrier, and the relationship between them. We choose the carrier to be an object or humanoid with which the human actor has potential contact.
We denote a sequence with $L$ frames as $\boldsymbol s=[\boldsymbol s^1, \boldsymbol s^2,\ldots, \boldsymbol s^{L}]$,
where $\boldsymbol s^i$ consists of human actor $\boldsymbol h^i$ and Carrier $\boldsymbol c^i$. Carrier refers to the object or the humanoid in the rest pose. Human actor $\boldsymbol h^i \in \mathbb{R}^{J \times D_h}$ is defined by $J$ joints with a $D_h$-dimensional representation at each joint, which is joint position and its velocity. Carrier $\boldsymbol c^i   \in \mathbb{R}^{N \times 3}$ is defined by $N$ object points or humanoid joints with its position.  
We denote the humanoid reactor sequence as $s_r=[\boldsymbol s_r^1, \boldsymbol s_r^2,\ldots, \boldsymbol s_r^{L}]$.
Given the sequence of human actor and carrier $s$, our goal is to predict a reasonable humanoid reactor sequence $s_r$.

\subsection{Social Affordance Representation} 

We define the social affordance representation centered on the carrier. Specifically, given a carrier, we encode the human actor motions to obtain a dense human-carrier representation. With this representation, we propose a social affordance representation that contains the motion of the human actor, the carrier’s dynamic geometry, and the actor-carrier relationship up until each time step. Note that the social affordance representation refers to the data stream from the beginning to a certain time step, rather than the representation of a single frame. 
The advantage of the social affordance representation is that it tightly associates the local region of the carrier with the human actor's motion, forming a strong representation for network learning.

\noindent\textbf{Carrier-centric actor representation.} Given a human actor $\boldsymbol h^i$ and a carrier $\boldsymbol c^i$ at time step $i$, we first define the carrier-centric actor representation $R^i$ as a collection of point-wise vectors on a set $\{x^i_j\}^N_{j=1}$ of N points or joints from carrier $\boldsymbol c$.
\begin{align}
    \boldsymbol R^i(\boldsymbol h^i, \boldsymbol c^i) = \{( \boldsymbol x^i_j, \boldsymbol \epsilon_{\theta}(h^i))\}_{j=1}^N,
\end{align}
where $R^i$ is carrier-centric actor representation, $x^i_j$ is the position of the point or joints from the carrier at time step $i$, $h^i$ is the human actor at time step $i$ and $\epsilon_{\theta}$ is the GNN network to encode human actor's motion to an embedding. 

\noindent\textbf{Social Affordance representation.} Given the carrier-centric actor representation at each time step, we define the social affordance representation $A^i$ as a collection of  $\{R^t\}^i_{t=1}$ up until each time step.

\begin{align}
    \boldsymbol A^i = \{\boldsymbol R^t\}^i_{t=1} = \{\{( \boldsymbol x^t_j, \boldsymbol \epsilon_{\theta}(h^t_j))\}_{j=1}^N\}^i_{t=1},
\end{align}
where $A^i$ refers to the social affordance representation at time step $i$ and $R^t$ refers to carrier-centric actor representation at time step $i$.

\subsection{Social Affordance Canonicalization}

\noindent\textbf{Learning Local Frames for Carrier.}
We believe that a local frame can reflect the geometric information of the carrier and can reflect the contact information between the carrier's local area and the human actor. 
Let $\boldsymbol c$ denote the carrier and $\{x_j\}^N_{j=1}$ as each point or joint. Let $H$ and $\textbf{V}$ denote per-point invariant scalar features and equivariant vector features of $\boldsymbol c$, respectively. 

We use $\boldsymbol c, H_{in},\textbf{V}_{in}$ to denote the carrier, invariant scalar, and equivariant vector features, where $H_{in},\textbf{V}_{in}$ are all zeros. We pass the carrier to an Equivariant network that aims to extract invariant and equivariant features, denoted as \text{EquivLayer}. Our \text{EquivLayer} is adapted from the GVP-GNN layer\cite{jing2020learning}.

\begin{equation}
(H_{out}, \textbf{V}_{out}) \gets \text{EquivLayer}(\boldsymbol c, H_{in},\textbf{V}_{in}).
\end{equation}
Since $\text{EquivLayer}$ is equivariant at all the layers, and the inputs $H_{in}, \textbf{V}_{in}$ are invariant and equivariant features, the output $H_{out},\textbf{V}_{out}$ of each layer are also invariant and equivariant features, respectively.

To obtain local frames of each point from invariant and equivariant features, we use another set of equivariant networks adapted from the GVP layers\cite{jing2020learning}. We use $\text{FrameNet}$ to denote the network.
\begin{align}
    \textbf{V}_{out} \gets \text{FrameNet}(H_{out}, \textbf{V}_{out}),
\end{align}
where each frames will be constructed from the equivariant features $\textbf{V}_{out}=(\textbf{v}_{out,1},\cdots,\textbf{v}_{out,N})(\textbf{v}_{out,j}\in\mathbb R^{2\times 3})$. 

We orthonormalize the two vectors $\textbf{v}_{out,j,1},\textbf{v}_{out,j,2}$ for each point to get $\textbf{u}_{j,1},\textbf{u}_{j,2}$ using the Gram-Schmidt method. Then we get the local frame $\textbf{F}_l= \{\textbf{F}_j\}_{j=1}^N=\{\textbf{u}_{j,1}, \textbf{u}_{j,2}, \textbf{u}_{j,1} \times \textbf{u}_{j,2} \}_{j=1}^N\in \mathbb R^{3\times 3}$. Since $\textbf{V}_{out}$ is rotation equivariant, the constructed frames are also rotation equivariant.
We refer to the whole module to generate local frames as \text{Equiv-FrameNet}:
\begin{align}
    \textbf{F}_l \gets \text{Equiv-FrameNet}(\boldsymbol c,H_{in},\textbf{V}_{in}),
\end{align}
where $\textbf{F}_l = \{\textbf{F}_j\}_{j=1}^N $ denotes the local frames of each point or joint from the carrier.

\noindent\textbf{Social Affordance Canonicalization.}
We propose a canonical social affordance canonicalization technique to simplify the distribution. We will explain in detail how to canonicalize social affordance using learned local frame $\textbf{F}_l$.

Since we have learned an equivariant local frame $\textbf{F}_l$ for every point or joint from the carrier, we first transform the motions of the human actor into the frame of each point or joint. Next, we encode the human actor's motions from the perspective of each point to obtain a dense object-centric HOI representation.  
This can be seen as binding an `observer' to each point on the object, and each `observer' encodes the actor's motions from a first-person view. The advantage is that it tightly associates the object's motion with the actor's motion, simplifying the distribution of social affordance and facilitating network learning.

\noindent\textbf{Canonical Carrier-centric actor representation.} Given a human actor $\boldsymbol h^i$ and a carrier $\boldsymbol c^i$ at time step $i$, we define the canonical carrier-centric actor representation $R_{c}^i$ as: 
\begin{align}
    \boldsymbol R_{c}^i(\boldsymbol h^i, \boldsymbol c^i) = \{( \boldsymbol x^i_j, \boldsymbol F^i_j, \boldsymbol \epsilon_{\theta}(h^i_j))\}_{j=1}^N,
\end{align}
where $F^i_j$ is the local frame, $h^i_j$ is the transformed human actor in frame $F^i_j$. 

\noindent\textbf{Canonical Social Affordance representation.} Based upon canonical carrier-centric actor representation $R_{c}^i$, we define the canonical social affordance representation $A_{c}^i$ as:

\begin{align}
    \boldsymbol A_{c}^i = \{\boldsymbol R_{c}^t\}^i_{t=1} = \{\{( \boldsymbol x^t_j, \boldsymbol F^t_j, \boldsymbol \epsilon_{\theta}(h^t_j))\}_{j=1}^N\}^i_{t=1},
\end{align}
where $A^i$ refers to the canonical social affordance representation at time step $i$ and $F^i_j$ is the local frame.

\subsection{Social Affordance Forecasting}

\begin{figure*}[t] 
	\centering  
	\includegraphics[width=\textwidth]{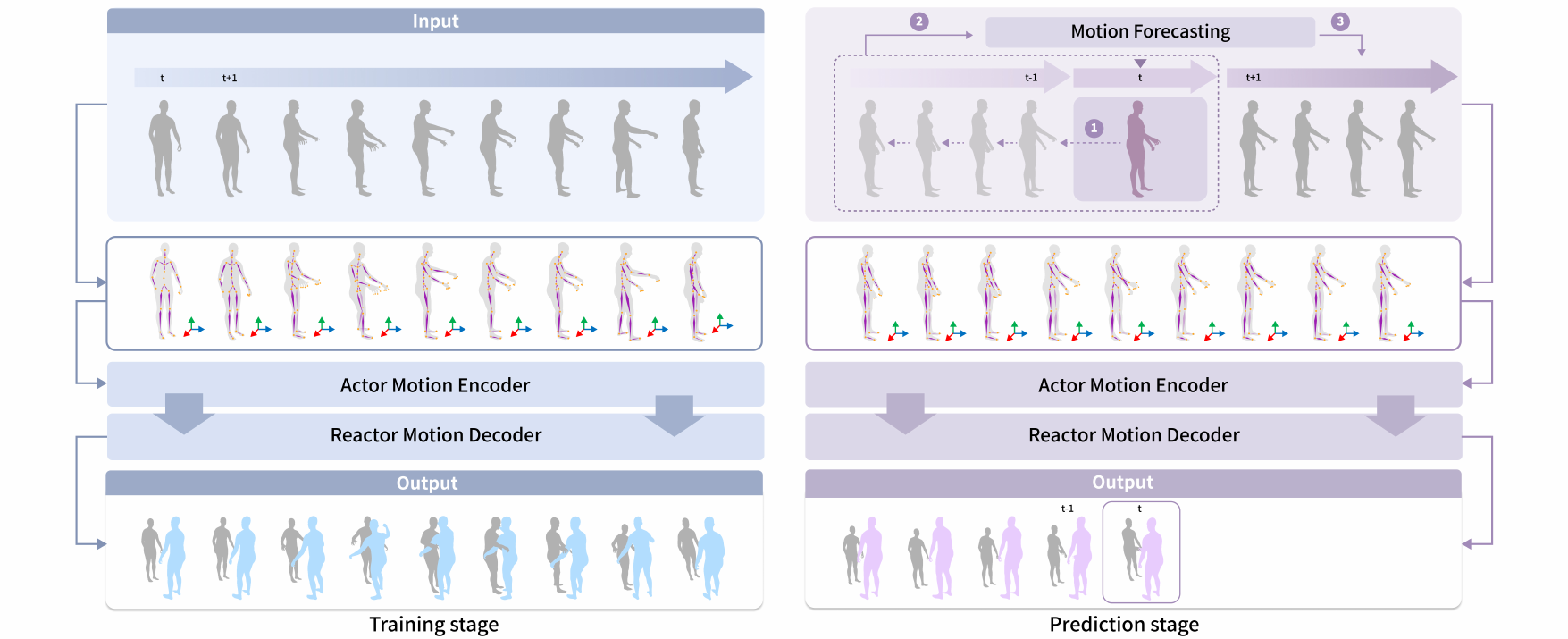} 
	\caption{Social Affordance Forecasting. At the training stage, the humanoid reactor can access all motions of the actor. At the prediction stage in the real world, the humanoid reactor can only observe the past motions of the human actor. The forecasting module can anticipate the motions that the human will take. } 
	\vspace{-7mm} 
    \label{fig:forecasting} 
\end{figure*}

Social affordance forecasting is to anticipate human actors' behavior so that the humanoid reactor can provide more reasonable responses. In real situations, the humanoid reactor can only observe the historical motions of the human actor. We believe that the humanoid reactor should have the ability to predict the motions of the human actor in order to better plan its own motions. For example, when someone raises their hand and walks towards you, you might instinctively think they are going to shake hands with you and be prepared for a handshake. Here, we introduce how to enable the reactor to make motion forecasting during prediction.

The $t$ observed motions of human actor and carrier are noted as  $\mathbf{s}^{ 1:M}=\left[ \mathbf{s}^{ 1 };\mathbf{s}^{ 2 };\ldots ; \mathbf{s}^{ M } \right]$. 
Therefore, the problem of online reaction synthesis is modeled as predicting$\mathbf{s_r}^{M}$ given $\mathbf{s}^{ 1:M }$. 
Given the observed motion $\mathbf{s}^{1:M}$, the objective of the motion forecasting problem is to predict the following $L$ motions $\mathbf{s}^{M+1: M+L}=\left[ \mathbf{s}^{M+1 };\mathbf{s}^{ M+2 };\ldots;\mathbf{s}^{ M+L } \right]$. 

We use a motion forecasting module to predict the human actor's motion and the object's motion(if available). For human-humanoid interaction setting, we use HumanMAC\cite{chen2023humanmac} as the forecasting module. For human-object-humanoid interaction setting, we build our motion forecasting module based on InterDiff\cite{xu2023interdiff} and add a prior that human-object contact is stable in order to simplify the difficulty in predicting the object's motion.
Finally, with the predicted result, we can obtain the  carrier-centric actor representation $\{R^i_{c}\}^{M+L}_i=1$, we define the canonical complete social affordance representation $A_{cf}^i$ as:

\begin{align}
    \boldsymbol A_{cf}^i = \{\boldsymbol R_{c}^t\}^{M+L}_{t=1} = \{\{( \boldsymbol x^t_j, \boldsymbol F^t_j, \boldsymbol \epsilon_{\theta}(h^t_j))\}_{j=1}^N\}^{M+L}_{t=1},
\end{align}
where $A_{cf}^i$ refers to the canonical complete social affordance representation at time step $i$ and $F^i_j$ is the local frame.

\subsection{Network and Objective Design}
Our network consists of a graph neural network and a 4D Transformer autoencoder. The graph neural network\cite{scarselli2008graph} transforms the human actor skeleton into a graph, efficiently modeling the relative motion between different joints. The 4D Transformer autoencoder\cite{wen2022point} is composed of a motion encoder and a motion decoder. The motion encoder takes the canonical complete social affordance as input and generates a latent embedding of it. The motion decoder uses the latent embedding as a condition and takes the previously taken motions of the reactor as input to generate new reaction motions autoregressively.

Specifically, given a sequence $s$, we can compute the $\boldsymbol A_{cf} $, and we can obtain the motions of the reactor $s_r$ by a 4D Transformer network.
\begin{align}
    \hat{s_r} = \textit{4DNet}(\boldsymbol A_{cf}) ,
\end{align}
where  $\textit{4DNet}$ denotes the whole 4D backbone to generate the motions of humanoid reactor.

We use two loss functions to train our model. The first one is the sequence loss which compares the generated position of joints with the ground truth using the Mean Square Error. The second one is the velocities of each joint.
\begin{equation}
\label{eq:loss1}
    Loss = MSE(s_r -\hat{s_r}) + MSE(ds_r-\hat{ds_r}),
\end{equation}
where $s_r$ refers to the GT position of each joint and $ds_r$ refers to the velocities of it.


\section{Experiments}

\begin{table*}[t] \small
	\centering
		\resizebox{0.95\linewidth}{!}{%
\begin{tabular}{@{}c| c c c c c c c c cccc@{} } 
\toprule
\multirow{2}*{Method} & \multicolumn{3}{c}{FID $\downarrow$} & \multicolumn{3}{c}{Diversity $\rightarrow$} & \multicolumn{3}{c}{Accuracy $\uparrow$}& \multicolumn{3}{c}{User Preference$\uparrow$} \\ \cmidrule(lr){2-4} \cmidrule(lr){5-7} \cmidrule(lr){8-10}  \cmidrule(lr){11-13}
 & HHI & InterHuman\cite{liang2023intergen} & Chi3D\cite{fieraru2020three} & HHI & InterHuman\cite{liang2023intergen} & Chi3D\cite{fieraru2020three} & HHI & InterHuman & Chi3D\cite{fieraru2020three} & HHI & InterHuman\cite{liang2023intergen} & Chi3D\cite{fieraru2020three}\\
\midrule
Real & 0.21 & 0.17 & 0.05 & 10.8 & 12.4 & 14.0 & 88.2 & - & 80.4 & - & -  &  - \\
\midrule
PGBIG\cite{ma2022progressively} & 56.7 & 87.2 & 67.2 & 13.9 & 17.1& 17.8& 34.1 &- & 61.6 & 4.4& 1.0&8.3 \\
SS-Transformer\cite{aksan2021spatio} & 77.8 & 107.3& 54.9& 16.2& 18.5& 19.2& 51.9 &- & 57.1& 2.7& 4.6& 18.4\\
InterFormer\cite{chopin2023interaction} & 54.3 & 73.1 & 20.8& 14.1 & 14.2& 14.8& 77.9 &- & 62.2 & 6.0& 2.1& 13.7\\
InterGen-Revised\cite{liang2023intergen} & 19.8 & 25.7 & 17.7& 11.6 & 13.3 & 14.2& 80.2 &- &71.9 & 19.7& 41.7&15.4 \\
\midrule
Ours & \textbf{13.3} & \textbf{14.7} & \textbf{12.8} & \textbf{11.1} & \textbf{13.3} & \textbf{14.1} & \textbf{85.4} & - & \textbf{77.6} & \textbf{67.2} & \textbf{50.6}& \textbf{44.2}\\
\bottomrule
\end{tabular}}
\caption{Quantitative results on HHI, InterHuman, and Chi3D. Our method consistently outperforms the previous method in all metrics.}
\label{tab:hhi}

\end{table*}

\begin{figure*}[th] 
	\centering  
	\includegraphics[width=1\textwidth]{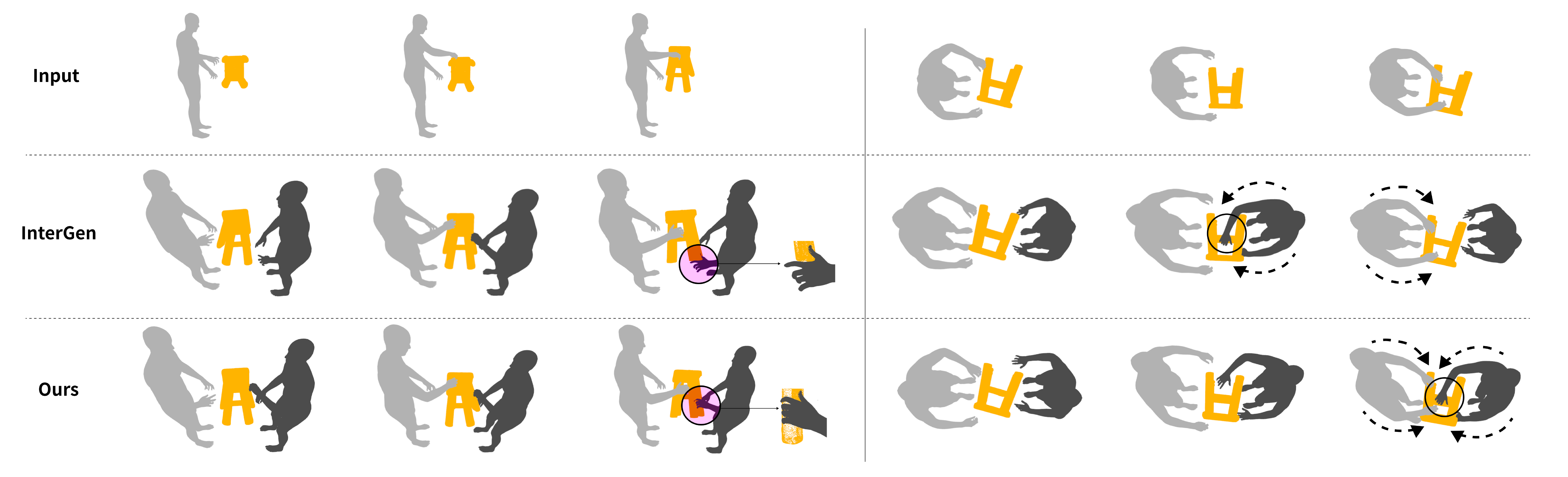} 
 	\vspace{-5mm} 
	\caption{Visualization results on CoChair. Our method can provide a more reasonable grasp and better collaboration with the human actor.} 
	\vspace{-8pt} 
    \label{fig:cochair} 
\end{figure*} 
\subsection{Datasets and Metric }
\noindent\textbf{CoChair Datasets.}
We use 2000 sequences of data as the training set and 1000 sequences as the test set. The one initiating the motion is the actor and he/she knows the carrying destination, while the other person is the reactor.

\noindent\textbf{HHI Datasets.}
We use 3000 sequences of data as the training set and 2000 sequences as the test set. The one initiating the action is the actor, while the other person is the reactor. It includes a total of 30 action categories.

\noindent\textbf{InterHuman Datasets.\cite{liang2023intergen}}
According to the official splits, we have selected 5200 sequences as the training set and 1177 sequences as the test set. Since there is no obvious initiator of actions in this dataset, we assume that the first person is the actor and the other person is the reactor. The dataset does not have fixed action categories, and each action corresponds to several textual descriptions. 

\noindent\textbf{Chi3D Datasets.\cite{fieraru2020three}}
We have a total of 373 available data provided by officials. Among them, 257 are used as the training set and 116 as the testing set. We consider the person estimated from images as actors and the other one captured by motion capture devices as reactors. 

\noindent\textbf{Metrics.} We use metrics commonly used in motion generation for quantitative results including action recognition accuracy, Frechet Inception Distance, and diversity.
Classification Accuracy measures how well our generated samples are classified by a motion classifier.
FID computes the distance between the ground truth and the generated data distribution.
Diversity Score is the average deep feature distance between all the samples.
For human-object-humanoid setting, we also report the mean penetration depth(cm)  when the distance between objects and generated reaction grasps is smaller than 0.2cm, which is commonly used for hand-object interaction\cite{zheng2023cams,zhang2021manipnet}. 
For motion feature extraction, we train an action recognition model using PPTr\cite{wen2022point} on each dataset and use it as the motion feature extractor. Due to CoChair and InterHuman not having a clear motion category, we use the feature extractor trained on the HHI dataset. PPTr is a Transformer-based 4D backbone to extract interaction motion features. We generate 1000 samples 10 times with different random seeds.

\begin{table}[t] \small
	\centering
		\resizebox{0.95\linewidth}{!}{%
\begin{tabular}{@{}c| c c c c@{} } 
\toprule
\multirow{1}*{Method} & \multicolumn{1}{c}{FID $\downarrow$} & \multicolumn{1}{c}{Diversity $\rightarrow$} &  \multicolumn{1}{c}{Penetration depth$\downarrow$} &  \multicolumn{1}{c}{User Preferenc$\uparrow$}\\
\midrule
Real &0.07 & 16.4 & 0.5& - \\
\midrule
PGBIG\cite{ma2022progressively} & 47.6 & 14.8& 7.2&3.5\\
SS-Transformer\cite{aksan2021spatio} & 51.2& 15.7& 3.7&3.3\\
InterFormer\cite{chopin2023interaction} & 44.2& 15.5& 4.2&6.4\\
InterGen-Revised\cite{liang2023intergen} & 26.7& 17.4& 2.2 &28.0\\
\midrule
Ours & \textbf{7.8} & \textbf{16.9}& \textbf{0.9} & \textbf{58.8}\\
\bottomrule
\end{tabular}}
\caption{Quantitative results on CoChair dataset. }
\label{tab:cochair}
\vspace{-7mm}
\end{table}

\subsection{Baselines}
For all baseline methods, we entirely used the author's code or made some modifications to adapt it to our task.

\noindent\textbf{Progressively Generating Better Initial Guesses}\cite{ma2022progressively} uses Spatial Dense Graph Convolutional Networks and Temporal Dense Graph Convolutional Networks. 

\noindent\textbf{Spatio-temporal Transformer}\cite{aksan2021spatio} is a transformer-based architecture that uses attention to find temporal and spatial correlations to predict human motion.

\noindent\textbf{InterFormer}\cite{chopin2023interaction}  consists of a Transformer network with both temporal and spatial attention to capturing the temporal and spatial dependencies of interactions. 

\noindent\textbf{InterGen-revised}\cite{liang2023intergen} is a powerful diffusion-based framework that can generate multi-human interaction based on the text description. 
We replace the CLIP branch with a spatio-temporal transformer to encode the actor's motion.

\begin{figure*}[th] 
	\centering  
	\includegraphics[width=1\textwidth]{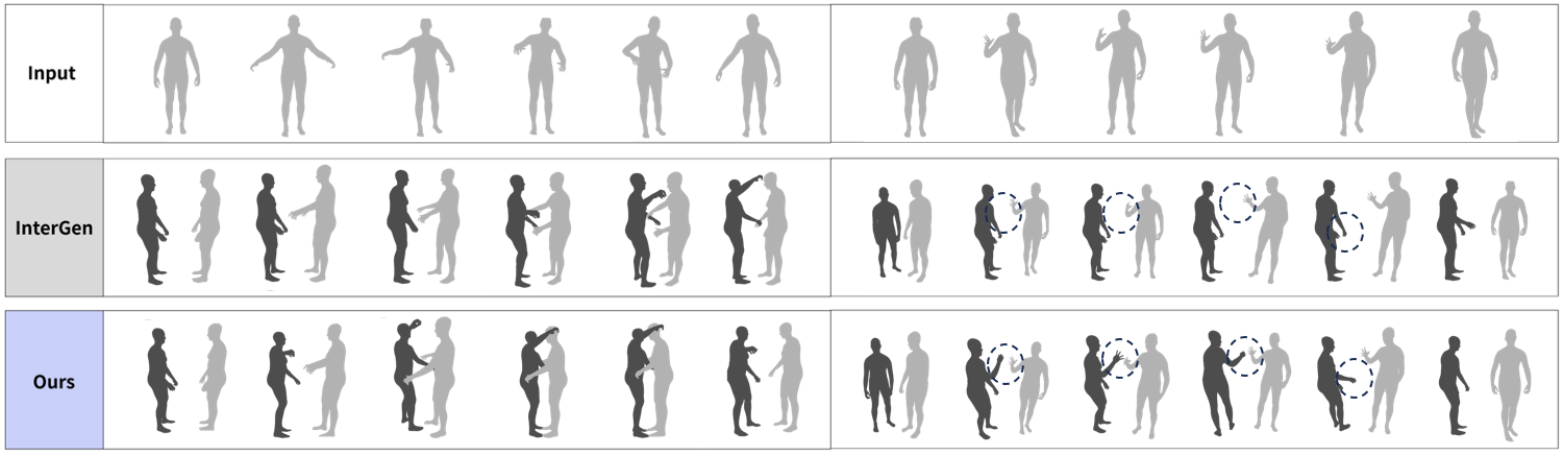} 
 \vspace{-5mm} 
	\caption{Visualization results on HHI. Our method can generate more prompt reactions and can better capture hand motion.} 
	\vspace{-8pt} 
    \label{fig:HHI} 
\end{figure*}

\begin{figure}[th] 
	\centering  
	\includegraphics[width=0.5\textwidth]{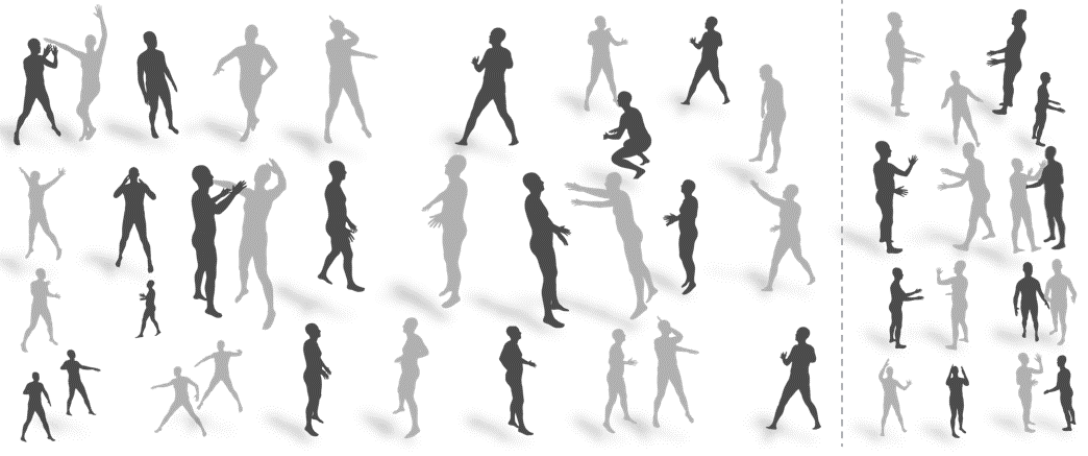} 
	\caption{Visualization gallery of our method on InterHuman (left) and Chi3D(right). The deep black one is generated by our method.} 
	\vspace{-8pt} 
    \label{fig:InterHuman} 
\end{figure} 

\subsection{Comparison to State-of-the-arts}
The results on the CoChair dataset are shown in Tab~\ref{tab:cochair}. Our method consistently outperforms previous method in all metrics. We compared the results with InterHuman and 
Fig~\ref{fig:cochair}, and it can be seen that our method can generate a more realistic and natural grasp(left) and collaboration(right). This indicates that through social affordance canonicalization, our approach can simplify the feature space, thus generating more complex and delicate motions. Through social affordance forecasting, we can anticipate the motions of human actors, thereby better planning cooperation with humans.
As for the human-humanoid interaction setting, the results are shown in Tab~\ref{tab:hhi}, our method outperforms baselines in all metrics. Some visualization results are shown in Fig~\ref{fig:HHI}. Compared with InterGen, our method can generate prompt reaction(left), and can better capture the local hand motions(right), while InterGen fails. We show that our method still can outperform the state-of-the-art method on existing human-human interaction datasets. We show the visualization results on InterHuman and Chi3D as shown in Fig~\ref{fig:InterHuman}, results demonstrate that our method generates a wide variety of realistic reactions.


\subsection{Ablation and Discussion}
\noindent\textbf{Ablation Study.} To validate our method, we conducted ablative experiments on the HHI dataset to verify the effectiveness of each design. Without canonicalization, our method drops significantly, indicating that the use of social affordance canonicalization to simplify feature space complexity is essential. Without social affordance forecasting, our method lost the ability to predict human actor motions, also leading to a performance drop. To verify the necessity of using the local frame, we also compared the effect of using a global frame, and it can be seen that our method is significantly superior. This also indicates that using a local frame to describe local geometry and potential contact is valuable.

\begin{table}[h] \small
	\centering
		\resizebox{0.95\linewidth}{!}{%
\begin{tabular}{@{}c| c c c c@{} } 
\toprule
\multirow{1}*{Method} & \multicolumn{1}{c}{FID $\downarrow$} & \multicolumn{1}{c}{Diversity $\rightarrow$} &  \multicolumn{1}{c}{Accuracy $\uparrow$}&  \multicolumn{1}{c}{User Preference$\uparrow$}\\
\midrule
Real & 0.21 & 10.8& 88.2 & - \\
\midrule
wo canonicalization &34.5&12.5& 78.4&13.4\\
wo forecasting & 16.7& 11.4& 82.1&19.4\\
w global frame &28.4& 8.9& 79.6&20.1\\
\midrule
Ours & \textbf{13.3} & \textbf{11.1} & \textbf{85.4} & \textbf{47.1}\\
\bottomrule
\end{tabular}}
\caption{Ablation study to justify each design of our method.}
\vspace{-5mm}
\label{tab:ablation}

\end{table}

\noindent\textbf{Visualization of Learned Local Frame for Carrier.} We visualized the frames on the rest-posed humanoid carrier. It can be seen that the frames on the spine are basically consistent, and the joint frames on the left and right sides exhibit roughly symmetrical characteristics. We also visualize the local frames, and it can be seen that the frames on the chair legs are approximately the same and can be generalized to different chairs. 
\begin{figure}[th] 
	\centering  
	\includegraphics[width=0.4\textwidth]{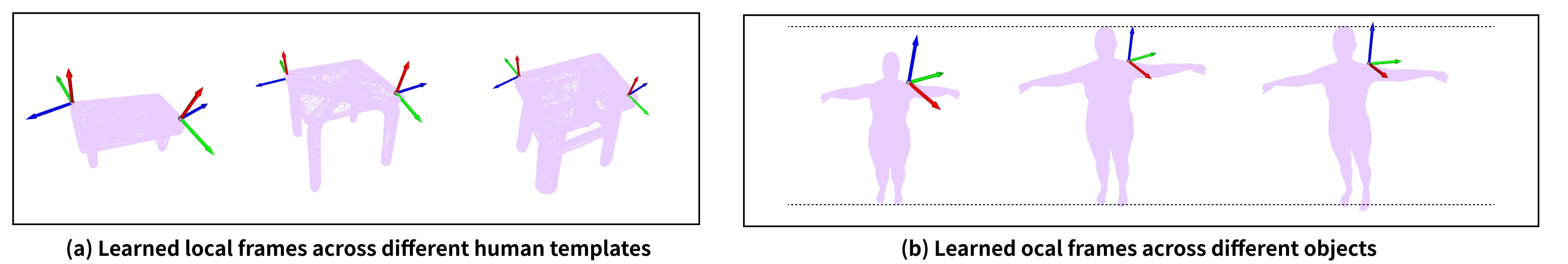} 
	\caption{Visualization results of learned local frame. The local frames are roughly consistent across different chairs.} 
	\vspace{-5mm} 
    \label{fig:frame} 
\end{figure} 

\noindent\textbf{Visualization of Motion Forecasting.}
We visualized the results of motion forecasting, and it can be seen that our method can provide reasonable predictions of whether there are objects or not. The imagination of these future behaviors can help the humanoid reactor to give more prompt and reasonable reactions.
\begin{figure}[th] 
	\centering  
	\includegraphics[width=0.5\textwidth]{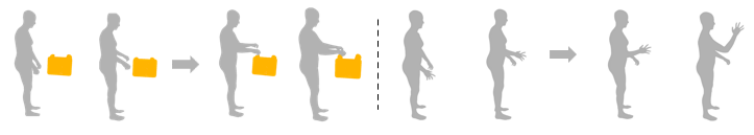} 
	\caption{Visualization results of Motion Forecasting with object(left) and without object(right).} 
    \label{fig:visforecast} 
\end{figure}

\noindent\textbf{Computation Overhead.}
We computed the computational overhead of our method and compared it with InterGen\cite{liang2023intergen}. All tests were conducted on a single 80G A100 graphics card. We set the input action sequence to 250 frames.  In terms of the number of parameters, our method is significantly more lightweight. The proposed social affordance canonicalization simplifies the distribution of actor motions, thus reducing the need for a large number of network parameters. Moreover, our method nearly achieves real-time inference at approximately 25 fps, whereas InterGen only reaches 0.54 fps. It's worth mentioning that our proposed FrameNet, utilizing merely 122 B of parameters, accomplishes the canonicalization of social affordance. Results show that our method has great potential and application in the fields of AR/VR games and humanoid robot control. 

\begin{table}[h]
    \centering
    \begin{tabular}{cccc}
    \toprule
         Method&  Memory  & Parameter & FPS\\
    \midrule
         InterGen-Revised\cite{liang2023intergen}& 38.12G &291.29M& 0.54fps\\
         Ours&  \textbf{22.28G} & \textbf{11.70M}& \textbf{25fps}\\
    \bottomrule
    \end{tabular}
    \caption{Our method is significantly more lightweight and can achieve real-time inference at approximately 25 FPS.}
    \label{tab:my_label}
    \vspace{-5mm}
\end{table}

\noindent\textbf{The Quality of Our Datasets.}
Our motion capture system has a high accuracy of 0.019 mm, effectively capturing contact regions as demonstrated in the figures below. We have thoroughly inspected the dataset to ensure its quality.
\begin{figure}
    \centering
    \includegraphics[width=1\linewidth]{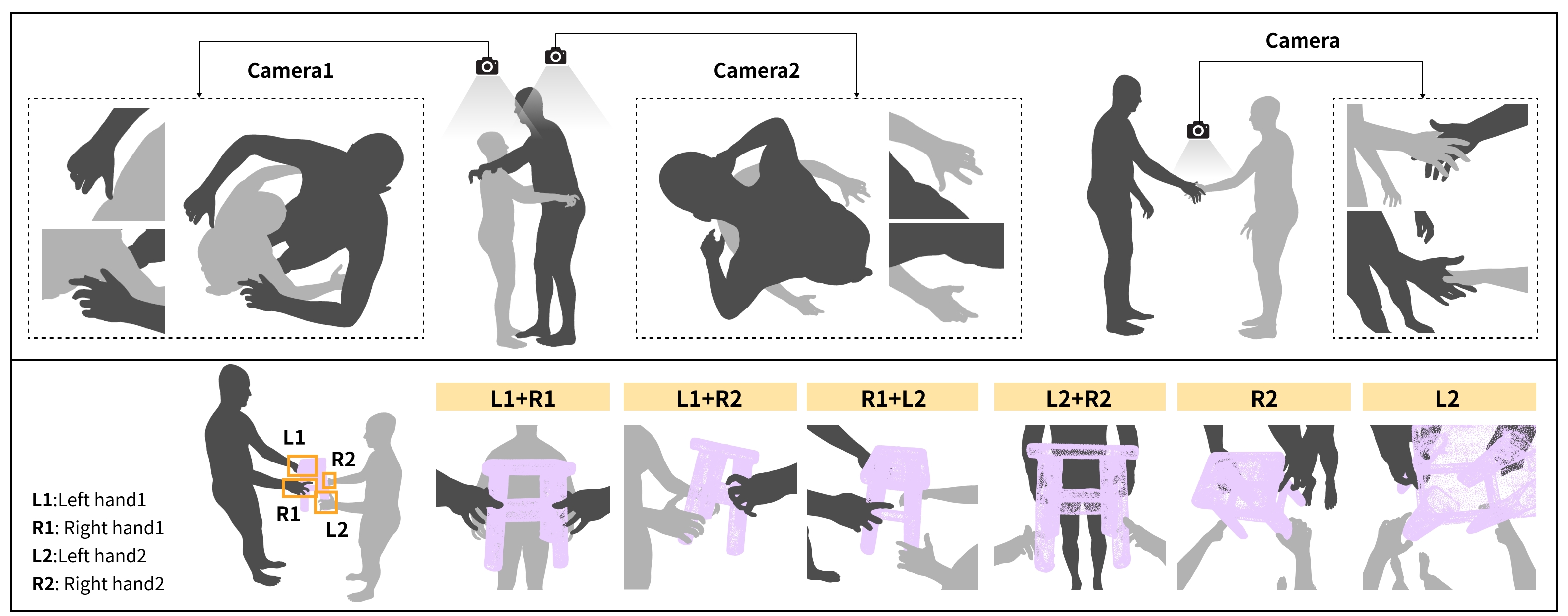}
        	\vspace{-5mm} 
    \caption{The Quality of Our Datasets.}
    \label{fig:enter-label}
    	\vspace{-8mm} 
\end{figure}

\section{Conclusions}
We propose a new task named online full-body motion reaction synthesis, which generates humanoid reactions based on the human actor's motions. We construct two datasets to support the research on the reaction synthesis task, HHI for Human-Humanoind Interaction and CoChair for Human-Object-Humanoind Interaction. We propose a novel technique named social affordance canonicalization and forecasting to achieve realistic and natural humanoid reaction synthesis. Experiments demonstrate that our approach can effectively generate high-quality reactions on proposed datasets and existing datasets. 



{
    \small
    \bibliographystyle{ieeenat_fullname}
    \bibliography{main}
}


\end{document}